\documentclass{article} 
\usepackage{url}
\usepackage{nips13submit_e,times}

\usepackage{natbib}

\usepackage[dvips]{graphicx} 
\usepackage{epsfig} 

\usepackage{algorithm}
\usepackage{algorithmic}

\usepackage{lipsum,amsmath,multicol}
\usepackage{amssymb}

\title{Feature Graph Architectures}

\author{
Richard Davis\\
\texttt{richard.davis@sydney.edu.au} \\
\AND
Philip Leong \\
\texttt{philip.leong@sydney.edu.au} \\
Computer Engineering Laboratory \\
Information and Electrical Engineering \\
Building J03, Maze Cresent \\
University of Sydney \\
Australia 2006 \\
Tel.: +61 2 9351 4491
\AND
Sanjay Chawla \\
School of Information Technologies\\
Building J12, University of Sydney \\
Australia 2006 \\
Tel.: +61 2 9351 3516
\texttt{chawla@it.usyd.edu.au} \\
}

\newtheorem{theorem}{Theorem}

\newtheorem{definition}{Definition}

\graphicspath{{/}}

\begin{document}

\maketitle

\begin{abstract}
In this article we propose feature graph architectures (FGA), which are deep learning systems employing a structured initialisation and training method based on a feature graph which facilitates improved generalisation performance compared with a standard shallow architecture. The goal is to explore alternative perspectives on the problem of deep network training. We evaluate FGA performance for deep SVMs on some experimental datasets, and show how generalisation and stability results may be derived for these models. We describe the effect of permutations on the model accuracy, and give a criterion for the optimal permutation in terms of feature correlations. The experimental results show that the algorithm produces robust and significant test set improvements over a standard shallow SVM training method for a range of datasets. These gains are achieved with a moderate increase in time complexity.
\end{abstract}

\section{Introduction}

In a recent review, Bengio \cite{DeepLearningOfRepresentations} emphasised that deep learning algorithms which can do a much better job of disentangling the underlying factors of variation in machine learning problems would have tremendous impact. Towards this goal we propose feature graph architectures (FGAs), which are hierarchical networks of learning modules (such as support vector machines) with some advantageous properties.
FGA is a deep learning model, similar to deep belief networks \cite{Hinton1} and deep support vector machines \cite{DSVM}. The training method involves three new components. It partitions the features into a subspace tree architecture, initialises the structure to give the same output as an optimised shallow SVM, and then identifies the subspaces at each part of the structure in which improved generalisation can be found. This method ensures that any changes that are found can only improve the training set error, and within certain generalisation bounds, the test error will improve with high probability also.

The key motivation behind this work is that the method is able to generate significant improvements in test set accuracy compared with a standard SVM using improved fits in the parameter subspaces, exploiting the SVM initialisation and selective node training to improve performance over a standard shallow SVM. The method consistently achieves significantly improved test error performance over standard SVM models in practise. We provide experimental evidence verifying the improved training and test performance of FGA on a range of standard UCI datasets \cite{UCIdatasetref} as well as synthetic problems. A derivation of a generalisation bound is provided to illustrate how this type of analysis may be performed for the FGA. We analyse the stability of the FGA under changes in the training data and demonstrate that there is a clear tradeoff between generalisation accuracy and stability in these models. We also investigate the dependence of the FGA on permutations of the input nodes. Permutation of the inputs to ensure decorrelation in the final layers enables improved fitting and generalisation. Theoretical stability and correlation analysis results are supported by numerical experiments, and an analysis of the time complexity of the FGA algorithm is given.

\section{Existing work on SVM architectures}
\label{sec:1}
Support Vector Machines have been applied to a wide range of applications, and there are many variants. These can be differentiated into linear and nonlinear, regression and classification, and a range of alternatives for the kernel function. See \cite{SVMoverview} for an overview, and \cite{SVMsoft} for an extensive list of widely used packages covering the major types of SVM implementations. SVM complexity and generalisation are discussed in \cite{Burges98atutorial, MohriBook}. For sparse, high-dimensional data,  linear support vector machines have proven effective \cite{Thorsten}. \cite{bottou2007large} gives an overview of scaling SVMs for large datasets.

There have been a number of attempts at organising SVMs into hierarchical structures \cite{Diez:2010:SDA:1837515.1837558}. SVMs have been combined with Self-Organising Maps \cite{SOMSVMtreesCao} which first divide the dataset into subsets, to which SVMs are applied separately. Support Vector based clustering was investigated by \cite{Horn}.

SVMs have been used to implement decision trees by placing an SVM at each node of the tree \cite{Bennett,gjorgji2}. Ensemble methods such as bagging and random forests have been applied to binary SVM decision trees \cite{Gjorgji}. Hierarchical SVM networks have been constructed by several authors by dividing the dataset into subsets using SVMs for clustering \cite{SVMcluster}. 
Chen and Crawford  \citep{Chen04integratingsupport} organised SVMs into hierarchical classification architectures, where the output classes were subdivided in a hierarchical manner.

Deep learning using support vector machines (deep-SVM, or DSVM) has been investigated by several researchers. In \cite{Abdullah} an ensemble of DSVMs was applied to image categorisation. Recently, backpropagation training methods for DSVM were given in \cite{DSVM} which were shown to often, but not always, provide some generalisation performance improvements compared with a simple SVM.



In this paper we propose and study the generalisation performance of feature graph architectures (FGA). We also investigate the stability of the FGA predictor, and the effect of permuting the inputs on generalisation error. The method is novel due to its organisation of the features into a specific subspace tree architecture, the initialization using the coefficients of a shallow model (in our case, SVM), and the selective training of each node to the target, scaled appropriately at each node. The key reason for this architecture is to exploit the structure of subspaces of the full feature space in order to achieve improved generalization performance. The relevant calculations are given in a series of derivations in the appendices, and described in the following sections.


\section{Feature Graph Architectures}
We consider a distribution $\mathcal{D}$ over $\mathcal{X} \times \mathcal{Y}$ where $\mathcal{X}$ is the feature space of dimension $d$ and $\mathcal{Y}\in \mathbb{R}$ is the target space. We define the expected loss or generalisation error $R(h)$ with respect to the function $h$ as the expectation
\begin{align}
R(h)=\underset{x\sim D}{E}\left[L(h(x),f(x))\right]
\end{align}
and the empirical loss or error of $h\in H$ for a sample of points $S=\left\{(x_i,y_i) : i=1 \ldots m\right\}$ of size $m$ from $\mathcal{D}$ as the mean
\begin{align}
\hat{R}(h)=\frac{1}{m}\sum_{i=1}^m L(h(x_i),y_i)
\end{align}

\label{sec:2}
\begin{figure}[ht]
\centering
\includegraphics[width=8cm]{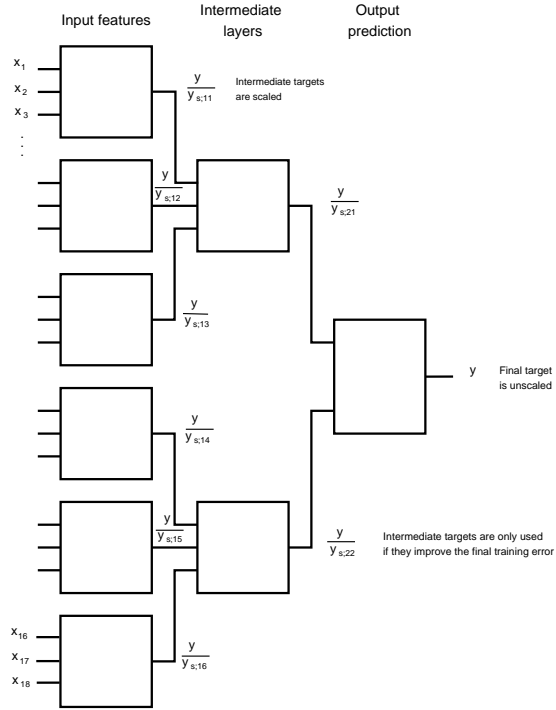}
\caption{Loss-optimised feature graph architecture\label{fig:layered}. After initialising to a shallow SVM, each node is trained to a common target, $y$, with additional scaling $y_{s;m}$ for node $m$ and the predictions of each node on the training set are used as features for training subsequent nodes.}
\end{figure}

A feature graph is a lattice of machine learning modules, organized into a tree structure. In this article we focus on support vector machines, and also briefly touch on a neural network implementation. A potentially large set of input features is partitioned into small subsets and fed into the first layer of the tree. Each node is trained to the same target using an algorithm such as SVM. The model is then reapplied to the training set to generate outputs which are used in subsequent nodes in the lattice. The feature grouping can be selected either by iterating over multiple groupings, which can be time consuming, or alternatively a heuristic such as grouping to maximise feature correlation may be used. In a later section we provide justification for this. The optimal heuristic for different problems is an open (and non-trivial) question, but from our experiments this heuristic is very effective in practice.

The feature graph algorithm may be implemented in several forms, depending upon the required architecture. The simplest training method is the layer-based feature graph, which we include as a reference when studying the performance of the main algorithm, the loss-optimised FGA. The layer-based FGA algorithm is as follows:

{\it \vspace{5pt} Layer-based feature graph\newline }

Given the empirical error function $\hat{R}$,  form $L$ successive layers of SVM models (see figure \ref{fig:layered}). There are $M_l$ nodes in layer $l$. For all layers:
\begin{enumerate}
\item Form feature groups of size $M$ in each layer and fit an SVM to those features using the common target $y$
\item Apply the SVM to those features and use the predictions as features in the next layer
\item Repeat the process for each successive layer until the final layer containing only one SVM is reached.
\end{enumerate}
The method is detailed in algorithm 1.
\begin{algorithm}
\caption{Layer-based FGA training algorithm}
\begin{algorithmic}
\REQUIRE $x_i \in \mathbb{R}^D, y_i \in \mathbb{R}, M \geq 2, \epsilon > 0, L, \{M_l\}  $
\WHILE{$\delta \hat{R} > \epsilon$}
\FOR{$l = 1$ to $L$}
\FOR{$m = 1$ to $M_l$}
  \STATE ${\rm FGA}[l;m] \leftarrow {\rm svm.train}(x_{l;m},y)$
  \STATE $\delta \hat{R} \leftarrow ||{\rm FGA.evaluate}(x) - y||$
\ENDFOR
\ENDFOR
\ENDWHILE
\end{algorithmic}
\label{layerfga}
\end{algorithm}

Our main focus will be the loss-optimised FGA, which decorrelates input variables in the early layers of the FGA, initialises the FGA to the coefficients of a simple SVM, and then selectively trains nodes in the graph. This algorithm is used in the generalisation and stability analysis later in the manuscript. The algorithm is as follows.

{\it \vspace{5pt} Loss-optimised Feature Graph Architectures\newline}

In the loss-optimised FGA, we first initialise the FGA with the parameters of the tuned SVM in the first layer, and set the weights and intercepts to $w=1, b=0$ respectively in subsequent layers. This will guarantee that its output exactly matches that of the standard SVM with linear kernels. We then iterate through the architecture, training each local node to a scaled copy of the target $y$. The scaling preserves the mean output of the node being trained, so that retraining a node produces an output of a scale suitable for the next downstream node in its current state. Only local modifications to a node which improve the overall $l^{2}$-error accuracy on the training set are retained. The steps are detailed in algorithm 2. 
Any parameter adjustments which generate improvements in a local cost function are verified against the global $l^2$ metric before being retained. As in the layered case, a range of input vector permutations should be used to determine the best architecture. This guarantees an improved training error.

\begin{algorithm}[h!]
\caption{Loss-optimised FGA training algorithm}
\begin{algorithmic}
\REQUIRE $x_i \in \mathcal{X}, y_i \in \mathcal{Y}, \epsilon > 0, M \geq 2, L, \{M_l\}$
\STATE ${\rm FGA_{l=1}} \leftarrow {\rm SVM}_{\rm tuned}$
\STATE ${\rm FGA_{l>1}} \leftarrow w=1,b=0$
\STATE $\delta \hat{R}_{\rm best} \leftarrow \infty$
\WHILE{$\delta \hat{R} > \epsilon$}
\FOR{$l = 1$ to $L$}
\FOR{$m = 1$ to $M_l$}
  \STATE ${\rm previous.node} \leftarrow {\rm FGA}[l;m]$
  \STATE $y_s \leftarrow y\frac{\bar{y}_{l;m}}{\bar{y}}$
  \STATE ${\rm FGA}[l;m] \leftarrow {\rm svm.train}(x_{l;m},y_s)$
  \STATE $\delta \hat{R} \leftarrow ||{\rm FGA.evaluate}(x) - y||$
    \IF {$\delta \hat{R} > \delta \hat{R}_{\rm best}$}
    \STATE ${\rm FGA}[l;m] \leftarrow {\rm previous.node}$
    \ELSE
    \STATE $\delta \hat{R}_{\rm best} \leftarrow \delta E$
    \ENDIF
\ENDFOR
\ENDFOR
\ENDWHILE
\end{algorithmic}
\label{l2FGA}
\end{algorithm}

{\it \vspace{5pt}Permutation optimised FGA\newline}
The loss-optimised FGA may be further refined by permuting the input set and selecting the permutation with the lowest training error. From the derivation in the next section, permutations which optimise the training error will improve the test error with high probability for sufficiently large $m$. In this algorithm we search over permutations to maximise the training error. In addition, in section 6 we show that the best permutation is one where FGA correlations in the early layers are higher, and correlations in later layers are lower (this was also confirmed numerically in our test).

\section{Generalisation bounds}
A range of generalisation bounds may be derived for a support vector regression problem. See \cite{MohriBook}, chapter 10 for details. Here we illustrate how such bounds may be used to derive associated bounds for a specific FGA, using the first bound of theorem 10.8, page 254 of \cite{MohriBook} as a starting point, since this is specifically designed for a single SVM. See appendix A for details of the derivation.

\begin{theorem}
Consider any node of the FGA with dimension $d\leq d_{\rm max}$, where $d_{\rm max}$ is the maximum dimensionality of each node over the FGA. Assume we have selected a fixed feature permutation using a heuristic. Let $K:X \times X \rightarrow R$ be a PDS kernel, let $\Phi : X \rightarrow H$ be a feature mapping associated to $K$ and let $H = \{ x \rightarrow w \cdot \Phi (x): ||w||_H \leq \Lambda\}$. Assume that there exists $r > 0$ such that $K(x,x)\leq r^2$ and $|f(x)| \leq \Lambda r$ for all $x \in X$. Fix $\epsilon >0$. $|\cdot|_{\epsilon}$ denotes the $\epsilon$-insensitive loss function (see \cite{MohriBook}, p252). Then, for any $\delta > 0$, for all $h\in H$ we have
\begin{equation}
R_{FGA}-R_{SVM} \leq \hat{R}_{FGA} - \hat{R}_{SVM} +|V| \epsilon + |V| \frac{2r\Lambda}{\sqrt{m}}\left(1+\sqrt{\frac{\log\frac{1}{\delta}}{2}}\right)
\label{genbound}
\end{equation}
with probability at least $1-|V|\delta$, where there are $|V|$ nodes in the FGA, and $\Lambda=d_{\max}$. This implies that the FGA has a lower generalization error than the corresponding SVM with high probability, for sufficiently large $m$, since the FGA training error is guaranteed to be less than (or if no node improvements are found, equal to) the SVM training error by construction.
\end{theorem}
Although the bound increases with $|V|$, to achieve the same probability we can use a smaller value of $\delta$, but $\delta$ appears inside the logarithmic term, and there is an additional square root. This gives an additional $\sqrt{\log |V|}$ dependence. This may seem onerous, but in practise we find that relatively few nodes are modified (in our tests it was usually around one in three nodes, resulting in a relatively small value of the $|V|\sqrt{\log{|V|}}$ dependence in the above bound, for constant values of $1-|V|\delta$.

This result applies to layer-based FGA, selective loss-optimised FGA, and permutation-optimised FGA, since the training error is improved and the FGA loss function is the same in each case. The proof is given in appendix A. Note that the bound improves if only $|V'| < |V|$ nodes yield training error improvements, since only those nodes are modified.

We can add equation \ref{genbound} to the standard SVM generalisation bound given in appendix A, equation \ref{standardsvmgenbound}, to obtain
\begin{equation}
R_{FGA} \leq \hat{R}_{FGA} + (|V|+1)\epsilon + (|V|+1) \frac{2r\Lambda}{\sqrt{m}}\left(1+\sqrt{\frac{\log\frac{1}{\delta}}{2}}\right)
\label{genboundfga}
\end{equation}
with probability at least $1-(|V|+1)\delta$, by the union bound.

\section{Stability}
Whilst the generalisation bound derived in the previous section does not require a stability analysis, we include it here to give the reader a sense of the tradeoff between stability and accuracy in the FGA. An algorithm is said to be uniformly stable \cite{BousquetElisseeff} if the following holds:
\begin{definition}{Uniform stability.} A learning algorithm $A:\cup_{m=1}^{\infty} Z^m \rightarrow F$ is said to be $\beta_m$-stable w.r.t. the loss function $L:\mathbb{R}\times \mathbb{R} \rightarrow \mathbb{R}$ if the following holds for all $i\in {1,\ldots,m}$.
\begin{align}
\exists z \in Z^m :\exists x \in Z: |L(f_z(x))-L(f_{z\backslash i}(x))|\leq \beta_m
\end{align}
\end{definition}

\begin{theorem}
An FGA is $\beta$-stable with stability given by
\begin{align}
\beta_{\rm FGA} = \beta_{\rm SVM} + \sum_{v \in V} \beta_v \prod_{i_v=v}^{v_L} w_{i_v}\label{stabilityeqn}
\end{align}
where $\beta_{\rm SVM}$ is the stability of a simple SVM, $\beta_v$ is the stability of an individual node $v$, and $w_{i_v}$ are the weights along paths from nodes $v \in V$  to the root $v_L$ of the FGA tree.
\end{theorem}
The proof is given in appendix B, together with some numerical examples. From this result it is apparent that the FGA has a lower stability than the corresponding simple SVM, although it has improved generalisation performance.

\section{Optimising Permutations}

Consider an arbitrary permutation of nodes.
Intuitively, the dominant contributions to the test error are correlations in the final layer. It is natural to minimise these in preference to early layer correlations, since their effect reaches over more input nodes. This means the optimal permutation is one where the correlation in the final layers is minimized and the correlation in the early layers is maximized. Any improvements in training and test errors in the FGA due to node permutations are governed by the generalisation bound of equation \ref{genbound}. For the following simplified situation we give an illustrative result to justify this heuristic.

{\bf Theorem 3.} Consider an FGA with four input features $1,2,3,4$. The second layer outputs are given by $X_{12},X_{34}$. A permutation of the input features gives second layer outputs $X_{13},X_{24}$.
If the input vectors are equally weighted, in the sense that $K=\langle Y,X_{12}\rangle=\langle Y,X_{34}\rangle=\langle Y,X_{13}\rangle=\langle Y,X_{24}\rangle$
 and $\langle X_{12},X_{34}\rangle\leq \langle X_{13},X_{24}\rangle$
 then the empirical errors for the two permutations satisfy
\begin{align}
R_{12,34} < R_{13,24}
\end{align}
{\bf Proof.}
The proof is shown in appendix C. From equation \ref{genbound} we see that training improvements imply test error improvements with high probability for sufficiently large training sets, so selecting permutations on the basis of training error and pairwise feature correlation is a reliable method for sufficiently large training sets.

However, in practise the situation is not always so simple, refer to the appendix for examples.

\section{Computational complexity}
For a space of $D$ dimensions, a standard SVM will scale as $O(D)$.  Since the FGA depth is $O(\log(D))$ and each node of dimension $m$ has complexity $O(m)$, a layer of size $d\leq D$ will have complexity $O(d)$. Thus the complexity of the corresponding FGA the complexity will be $O(D \log D)$, summing over all layers. This does not include searching over multiple input permutations, of which there are a very large number, but as mentioned the best results are achieved using a permutation which maximises the early-layer correlations and minimises later layer correlations, so in practise it is not necessary to do extensive permutation searches.

\section{Numerical results}
In this section we show the result of applying the FGA architecture by applying and comparing some widely used R packages \citep{Renv}: linear regression, NeuralNet \citep{neuralnetref}, RandomForest \citep{randomforestref} and support vector machines \citep{e1071}. Default parameters were used for the neural networks, with 5 nodes in the hidden layer. Default settings were used for the random forests run. The SVM used a linear kernel and tuned on the range $10^{(-10:10)}$.

The FGA architecture for a neural network is similar to the SVM. The output ranges of all nodes are scaled so that they lie in the non-asymptotic region of the following layer, ensuring that the network can be trained easily. The network is initialised using a single-layer network, with the second layer initialised so that it acts like an identity map, just as was done for the FGA-SVM.

It is not possible to cover all possible datasets, so we give results for one synthetic and four UCI datasets to illustrate the method.

\subsection{Regression, neural network and PCA comparison}
We compared standard and FGA approaches on a synthetic datasets corresponding to the equation
\begin{align}
y = \left(\sum_{i=1}^D x_i\right)^p
\end{align}
for a power $p=2$. The dataset dimension was $D=25$, with 5 features per node in the SVM. Figure \ref{svmfigs} shows the greatly enhanced accuracy of the FGA SVM on the test set. The results for simple and FGA architectures on linear regression, PCA-SVM (principle components analysis on the features), neural networks, FGA using neural network nodes (FGA-NN), random forests (RF), support vector machines and FGA using support vector machines (FGA-SVM) are shown in table \ref{firsttable}.

\begin{table}[h]\
\centering
\begin{tabular}{c c}
\hline
Model & $L=l^2$ error\\
\hline
 LR & 1.508\\
 PCA-SVM  & 1.177\\
 NN (2 layer) & 0.959 \\
 FGA-NN (2-layer)  & 0.754 \\
 RF & 0.753 \\
 SVM & 0.537 \\
 FGA-SVM L-opt., one perm. only  &  0.333\\
{\bf  FGA-SVM L-opt., optimised perm.}  & {\bf 0.323}\\
\hline
\end{tabular}
   \caption{$L=l^2$ error results using simple linear regresion (LR), neural networks (NN), FGA-neural networks (FGA-NN), random forests (RF), support vector machines (SVM), PCA support vector machines (PCA-SVM), and FGA support vector machines (FGA-SVM).\label{firsttable}}
\end{table}

\subsection{FGA-SVM comparison on some standard regression datasets}

To illustrate the effectiveness of the method on some standard datasets of realistic size, figure \ref{fig:allresultskernel} gives a summary of the results comparing SVM, loss-optimised FGA-SVM, and loss-optimised FGA-SVM with optimised permutations. We see that the test error is never worse than for the SVM, and significant improvements of up to 43\% were observed, depending on the dataset. Initialising to a permutation in which early layer nodes are highly correlated and later layer nodes are less correlated, followed by optimising over additional pairwise permutations, was able to provide significant additional improvements for the majority of datasets we tried. For the four sets presented here, this optimisation was very effective for two cases, and had negligible effect in the other two cases. In one of those two cases the FGA was not able to find significant improvements over the standard SVM and so the permutation had no effect. In the other case the permutation resulted in significant training improvements with only a very small test error increase, so we find no significant generalisation benefit from this permutation change. However the strong performance of the method for other datasets would justify evaluating permutations.
\begin{figure*}[h!t]
\small
\centering
\begin{tabular}{lllllllllllll}
\hline\noalign{\smallskip}
 Dataset &Method&Kernel  &$N_{train}$&$N_{test}$&D&     $\hat{R}$&   $R$:  \vspace{1pt} \\
\hline
\noalign{\smallskip}\hline\noalign{\smallskip}
Triazines & Tuned Simple SVM &Linear & 100 & 60 &60 &342.3  &977.1  \\
 & Loss-optimised Tuned FGA-SVM & &  &  & &  318.3& {\bf 745.1} \\
 & Loss-opt. max 50 rand. perm.
& & &  & &  279.18& 749.2 \\
\noalign{\smallskip}\hline\noalign{\smallskip}
Space GA  & Tuned Simple SVM &Linear & 500 & 3200 & 6 &567,594 & 537,717  \\
 & Loss-optimised Tuned FGA-SVM & & &  & &567,594 &  537,717 \\
 & Loss-opt. max 50 rand. perm.
& & &  & &566,429 & {\bf 536,642} \\
\noalign{\smallskip}\hline\noalign{\smallskip}
Pyrim  & Tuned Simple SVM &Linear & 40 & 34 &27  & 31.93&179.3  \\
 & Loss-optimised Tuned FGA-SVM & & &  &  &24.04 & 177.7 \\
 & Loss-opt. max 50 rand. perm.
& & &  &  &17.79 & {\bf 151.1} \\
\noalign{\smallskip}\hline\noalign{\smallskip}
Housing  & Tuned Simple SVM &Linear & 300 & 206 &13  & 321.2&6883   \\
 & Loss-optimised Tuned FGA-SVM & & &  &  &231.4 & 5204 \\
 & Loss-opt. max 50 rand. perm.
& & &  &  &198.3 & {\bf 3872} \\
\noalign{\smallskip}\hline
 \end{tabular}
 \caption{Train and test error results using simple SVM and the two loss-optimised FGA strategies \label{fig:allresultskernel} for a linear kernel. In all cases the cross-validation tuning ranges used were C: $2^{-2},2^{-1},2^{-0},2^{1},2^{2},2^{3},2^{4},2^{5}$. The $C$ parameter for the tuned SVM was used throughout the FGA architecture. We found significant improvements for Pyrim and Housing. The dimensionality of Space GA was only 6 which was why the FGA was not able to find significant improvements on the training or test sets. For Triazines, significant training set improvements were found but the test set error was only marginally worse, which shows that the FGA is not overfitting significantly relative to the SVM.}
 \end{figure*}

 Substituting the values in table \ref{fig:allresultskernel} into equation \ref{genbound} we see in all cases the bound is satisfied, providing experimental confirmation of the bound.

\section{Conclusions}
We have introduced the FGA architecture and described an algorithm incorporating several novel elements which enables effective and robust training of a deep SVM. We have also demonstrated effective performance when the deep SVM is replaced by a deep neural network.  We have shown that the FGA is able to provide significant improvements in test error over a shallow SVM, and in particular these improvements are robust, in that for all the datasets we tested it was never significantly worse, and was usually better.  We have given examples of how to derive generalisation bounds for the FGA, and illustrated the tradeoff between generalisation and stability in the FGA. We have described the effect of input permutations on the FGA accuracy, and given a criterion for a good permutation heuristic.
The FGA training architecture can be used to provide a smooth transition from a shallow network to a deep network, with improved test errors at each stage of the transition. It provides a practical and effective way to initialise and train a deep network. Whilst subject to generalisation bounds, we have demonstrated through a range of examples that the improvements can be very significant. As the neural network results appear promising on synthetic data, they and other deep learning implementations of the FGA appear promising. The FGA can easily be extended to non-linear kernels, and combined with other types of learning systems. These topics will be investigated in future work.

%


\bibliography{rdavisnips2013}

\begin{thebibliography}{24}
\providecommand{\natexlab}[1]{#1}
\providecommand{\url}[1]{\texttt{#1}}
\expandafter\ifx\csname urlstyle\endcsname\relax
  \providecommand{\doi}[1]{doi: #1}\else
  \providecommand{\doi}{doi: \begingroup \urlstyle{rm}\Url}\fi

\bibitem[Abdullah et~al.(2009)Abdullah, Veltkamp, and Wiering]{Abdullah}
Abdullah, Azizi, Veltkamp, Remco~C., and Wiering, Marco~A.
\newblock An ensemble of deep support vector machines for image categorization.
\newblock In \emph{Proceedings of the 2009 International Conference of Soft
  Computing and Pattern Recognition}, SOCPAR '09, pp.\  301--306, Washington,
  DC, USA, 2009. IEEE Computer Society.
\newblock ISBN 978-0-7695-3879-2.
\newblock \doi{10.1109/SoCPaR.2009.67}.
\newblock URL \url{http://dx.doi.org/10.1109/SoCPaR.2009.67}.

\bibitem[Bache \& Lichman(2013)Bache and Lichman]{UCIdatasetref}
Bache, K. and Lichman, M.
\newblock {UCI} machine learning repository, 2013.
\newblock URL \url{http://archive.ics.uci.edu/ml}.

\bibitem[Ben-Hur et~al.(2001)Ben-Hur, Horn, Siegelmann, and Vapnik]{Horn}
Ben-Hur, Asa, Horn, David, Siegelmann, Hava~T., and Vapnik, Vladimir.
\newblock Support vector clustering.
\newblock \emph{Journal of Machine Learning Research}, 2:\penalty0 125--137,
  2001.
\newblock URL
  \url{http://jmlr.csail.mit.edu/papers/volume2/horn01a/rev1/horn01ar1.pdf}.

\bibitem[Bengio(2013)]{DeepLearningOfRepresentations}
Bengio, Yoshua.
\newblock Deep learning of representations: Looking forward.
\newblock \emph{CoRR}, abs/1305.0445, 2013.

\bibitem[Bennett \& Blue(1997)Bennett and Blue]{Bennett}
Bennett, K.~P. and Blue, J.A.
\newblock A support vector machine approach to decision trees.
\newblock In \emph{Department of Mathematical Sciences Math Report No. 97-100,
  Rensselaer Polytechnic Institute}, pp.\  2396--2401, 1997.

\bibitem[Bottou(2007)]{bottou2007large}
Bottou, L.
\newblock \emph{Large-Scale Kernel Machines}.
\newblock Neural Information Processing Series. Mit Press, 2007.
\newblock ISBN 9780262026253.
\newblock URL \url{http://books.google.com.au/books?id=MDup2gE3BwgC}.

\bibitem[Bousquet \& Elisseeff(2002)Bousquet and Elisseeff]{BousquetElisseeff}
Bousquet, Olivier and Elisseeff, Andr{\'e}.
\newblock Stability and generalization.
\newblock \emph{J. Mach. Learn. Res.}, 2:\penalty0 499--526, March 2002.
\newblock ISSN 1532-4435.
\newblock \doi{10.1162/153244302760200704}.
\newblock URL \url{http://dx.doi.org/10.1162/153244302760200704}.

\bibitem[Burges(1998)]{Burges98atutorial}
Burges, Christopher~J.C.
\newblock A tutorial on support vector machines for pattern recognition.
\newblock \emph{Data Mining and Knowledge Discovery}, 2:\penalty0 121--167,
  1998.

\bibitem[Cao(2003)]{SOMSVMtreesCao}
Cao, Lijuan.
\newblock Support vector machines experts for time series forecasting.
\newblock \emph{Neurocomputing}, 51\penalty0 (0):\penalty0 321 -- 339, 2003.
\newblock ISSN 0925-2312.
\newblock \doi{10.1016/S0925-2312(02)00577-5}.
\newblock URL
  \url{http://www.sciencedirect.com/science/article/pii/S0925231202005775}.

\bibitem[Chen et~al.(2004)Chen, Crawford, and Ghosh]{Chen04integratingsupport}
Chen, Yangchi, Crawford, Melba~M., and Ghosh, Joydeep.
\newblock Integrating support vector machines in a hierarchical output
  decomposition framework.
\newblock In \emph{In 2004 International Geosci. and Remote Sens. Symposium},
  pp.\  949--953, 2004.

\bibitem[Cristianini \& Shawe-Taylor()Cristianini and
  Shawe-Taylor]{SVMoverview}
Cristianini, Nello and Shawe-Taylor, John.
\newblock \emph{{An Introduction to Support Vector Machines and Other
  Kernel-based Learning Methods}}.
\newblock Cambridge University Press, 1 edition, March .
\newblock ISBN 0521780195.

\bibitem[D\'{\i}ez et~al.(2010)D\'{\i}ez, del Coz, and
  Bahamonde]{Diez:2010:SDA:1837515.1837558}
D\'{\i}ez, J., del Coz, J.~J., and Bahamonde, A.
\newblock A semi-dependent decomposition approach to learn hierarchical
  classifiers.
\newblock \emph{Pattern Recogn.}, 43\penalty0 (11):\penalty0 3795--3804,
  November 2010.
\newblock ISSN 0031-3203.
\newblock \doi{10.1016/j.patcog.2010.06.001}.
\newblock URL \url{http://dx.doi.org/10.1016/j.patcog.2010.06.001}.

\bibitem[Dimitriadou et~al.(2010)Dimitriadou, Hornik, Leisch, Meyer, {}, and
  Weingessel]{e1071}
Dimitriadou, Evgenia, Hornik, Kurt, Leisch, Friedrich, Meyer, David, {}, and
  Weingessel, Andreas.
\newblock \emph{{e1071: Misc Functions of the Department of Statistics (e1071),
  TU Wien}}, 2010.
\newblock URL \url{http://CRAN.R-project.org/package=e1071}.

\bibitem[G\"{u}nther \& Fritsch()G\"{u}nther and Fritsch]{neuralnetref}
G\"{u}nther, Frauke and Fritsch, Stefan.
\newblock {neuralnet: Training of neural networks}.
\newblock \emph{The R Journal}, \penalty0 (1):\penalty0 30--38, June .

\bibitem[Hinton \& Osindero(2006)Hinton and Osindero]{Hinton1}
Hinton, Geoffrey~E. and Osindero, Simon.
\newblock A fast learning algorithm for deep belief nets.
\newblock \emph{Neural Computation}, 18:\penalty0 2006, 2006.

\bibitem[Ivanciuc(2005)]{SVMsoft}
Ivanciuc, Ovidiu.
\newblock Svm software collection, 2005.
\newblock URL \url{{http://support-vector-machines.org/SVM_soft.html}}.

\bibitem[Joachims(2006)]{Thorsten}
Joachims, Thorsten.
\newblock Training linear svms in linear time.
\newblock In \emph{Proceedings of the 12th ACM SIGKDD International Conference
  on Knowledge Discovery and Data Mining}, KDD '06, pp.\  217--226, New York,
  NY, USA, 2006. ACM.
\newblock ISBN 1-59593-339-5.
\newblock \doi{10.1145/1150402.1150429}.
\newblock URL \url{http://doi.acm.org/10.1145/1150402.1150429}.

\bibitem[Liaw \& Wiener(2002)Liaw and Wiener]{randomforestref}
Liaw, Andy and Wiener, Matthew.
\newblock Classification and regression by randomforest.
\newblock \emph{R News}, 2\penalty0 (3):\penalty0 18--22, 2002.
\newblock URL
  \url{{http://cran.r-project.org/web/packages/randomForest/randomForest.pdf}}.

\bibitem[M.A.~Wiering \& Schomaker(2013)M.A.~Wiering and Schomaker]{DSVM}
M.A.~Wiering, M.~Schutten, A. Millea A.~Meijster and Schomaker, L.R.B.
\newblock Deep support vector machines for regression problems.
\newblock \emph{Workshop on Advances in Regularization, Optimization, Kernel
  Methods, and Support Vector Machines: theory and application}, 2013.
\newblock URL
  \url{www.esat.kuleuven.be/sista/ROKS2013/files/abstracts/MAWiering.pdf‎}.

\bibitem[Madzarov \& Gjorgjevikj(2009)Madzarov and Gjorgjevikj]{gjorgji2}
Madzarov, G. and Gjorgjevikj, D.
\newblock Multi-class classification using support vector machines in decision
  tree architecture.
\newblock In \emph{EUROCON 2009, EUROCON '09. IEEE}, pp.\  288--295, 2009.
\newblock \doi{10.1109/EURCON.2009.5167645}.

\bibitem[Madzarov et~al.()Madzarov, Gjorgjevikj, and Chorbev]{Gjorgji}
Madzarov, Gjorgji, Gjorgjevikj, Dejan, and Chorbev, Ivan.
\newblock A multi-class svm classifier utilizing binary decision tree.
\newblock \emph{Informatica (Slovenia)}, \penalty0 (2):\penalty0 225--233.

\bibitem[Mohri et~al.(2012)Mohri, Rostamizadeh, and Talwalkar]{MohriBook}
Mohri, Mehryar, Rostamizadeh, Afshin, and Talwalkar, Ameet.
\newblock \emph{Foundations of Machine Learning}.
\newblock The MIT Press, 2012.
\newblock ISBN 026201825X, 9780262018258.

\bibitem[{R Development Core Team}(2012)]{Renv}
{R Development Core Team}.
\newblock \emph{R: A Language and Environment for Statistical Computing}.
\newblock R Foundation for Statistical Computing, Vienna, Austria, 2012.
\newblock URL \url{http://www.R-project.org/}.
\newblock {ISBN} 3-900051-07-0.

\bibitem[Yu et~al.(2005)Yu, Yang, Han, and Li]{SVMcluster}
Yu, Hwanjo, Yang, Jiong, Han, Jiawei, and Li, Xiaolei.
\newblock {Making SVMs Scalable to Large Data Sets using Hierarchical Cluster
  Indexing}.
\newblock \emph{Data Mining and Knowledge Discovery}, 11\penalty0 (3):\penalty0
  295--321, November 2005.
\newblock ISSN 1384-5810.
\newblock \doi{10.1007/s10618-005-0005-7}.
\newblock URL \url{http://dx.doi.org/10.1007/s10618-005-0005-7}.

\end{thebibliography}
\bibliographystyle{icml2014}
\newpage
\appendix
\section{Generalisation bounds on the FGA}
{\bf Theorem 1.}
We refer the reader to the first bound of theorem 10.8, page 254 in \cite{MohriBook}. Assume we have selected a fixed feature permutation for the FGA using a heuristic.

 Consider any node of the FGA with dimension $d\leq d_{\rm max}$, where $d_{\rm max}$ is the maximum dimensionality of each node over the FGA. Assume we have selected a fixed feature permutation using a heuristic. Let $K:X \times X \rightarrow R$ be a PDS kernel, let $\Phi : X \rightarrow H$ be a feature mapping associated to $K$ and let $H = \{ x \rightarrow w \cdot \Phi (x): ||w||_H \leq \Lambda\}$. Assume that there exists $r > 0$ such that $K(x,x)\leq r^2$ and $|f(x)| \leq \Lambda r$ for all $x \in X$. Fix $\epsilon >0$. Then, for any $\delta > 0$, for all $h\in H$ we have

\begin{equation*}
R_{FGA}-R_{SVM} \leq \hat{R}_{FGA} - \hat{R}_{SVM} + |V|\epsilon + |V| \frac{2r\Lambda}{\sqrt{m}}\left(1+\sqrt{\frac{\log\frac{1}{\delta}}{2}}\right)
\end{equation*}
with probability at least $1-|V|\delta$ by the union bound, where there are $|V|$ nodes in the FGA, and $\Lambda=d_{\max}$ is the maximum dimensionality of each node. This implies that the FGA has a lower generalization error than the corresponding SVM with high probability, for sufficiently large $m$, since the FGA training error is guaranteed to be less than (or if no node improvements are found, equal to) the SVM training error by construction.

{\bf Proof}. With probability $1-\delta$ the SVM satisfies \cite{MohriBook}
\begin{align}
|R_{SVM} - \hat{R}_{SVM}|_{\epsilon} \leq  \frac{2r\Lambda}{\sqrt{m}}\left(1+\sqrt{\frac{\log\frac{1}{\delta}}{2}}\right)\label{standardsvmgenbound}
\end{align}
so
\begin{align}
R_{SVM} - \hat{R}_{SVM} - \epsilon \leq  \frac{2r\Lambda}{\sqrt{m}}\left(1+\sqrt{\frac{\log\frac{1}{\delta}}{2}}\right)
\end{align}
and
\begin{align}
\hat{R}_{SVM} - R_{SVM} - \epsilon  \leq  \frac{2r\Lambda}{\sqrt{m}}\left(1+\sqrt{\frac{\log\frac{1}{\delta}}{2}}\right) \label{SVMeq}
\end{align}
Any FGA with linear kernels may be expanded into a simple linear function $y=w.x+b$.
Since the FGA acts on the same space with the same loss function, it satisfies the same bound, namely
\begin{align}
R_{FGA} - \hat{R}_{FGA} - \epsilon  \leq  \frac{2r\Lambda}{\sqrt{m}}\left(1+\sqrt{\frac{\log\frac{1}{\delta}}{2}}\right) \label{FGAeq}
\end{align}
and
\begin{align}
\hat{R}_{FGA} - R_{FGA}  - \epsilon \leq  \frac{2r\Lambda}{\sqrt{m}}\left(1+\sqrt{\frac{\log\frac{1}{\delta}}{2}}\right)
\end{align}

We can add equations \ref{SVMeq} and \ref{FGAeq} to give
\begin{equation}
R_{FGA}- \hat{R}_{FGA} - R_{SVM} + \hat{R}_{SVM} \leq 2\epsilon + 2 \frac{2r\Lambda}{\sqrt{m}}\left(1+\sqrt{\frac{\log\frac{1}{\delta}}{2}}\right)
\end{equation}
with probability at least $1-2\delta$, for a single node retraining, by the union bound. Next, note that an identical equation can be written for a pair of FGAs which differ by a single node retraining. We can therefore sum over all node trainings (of which there are at most $|V|$, where every node has at most $d_{\max}$ features (as $|\mathcal{H}|=d_{\max}$ for an SVM with $d_{max}$ features per node), to obtain
\begin{equation}
R_{FGA}-R_{SVM}  \leq \hat{R}_{FGA} - \hat{R}_{SVM} + |V|\epsilon + |V| \frac{2r\Lambda}{\sqrt{m}}\left(1+\sqrt{\frac{\log\frac{1}{\delta}}{2}}\right)
\end{equation}
with probability at least $1-|\mathcal{V}|\delta$, by the union bound. Thus if we improve the training error, the test error will also improve with probability at least $1-|\mathcal{V}|\delta$ for sufficiently large $m$. Since the FGA only modifies nodes which improve the global training error, the test error must improve with high probability within the above bound.$\Box$.

\begin{figure*}[h!t]
\small
\centering
\includegraphics[width=9cm]{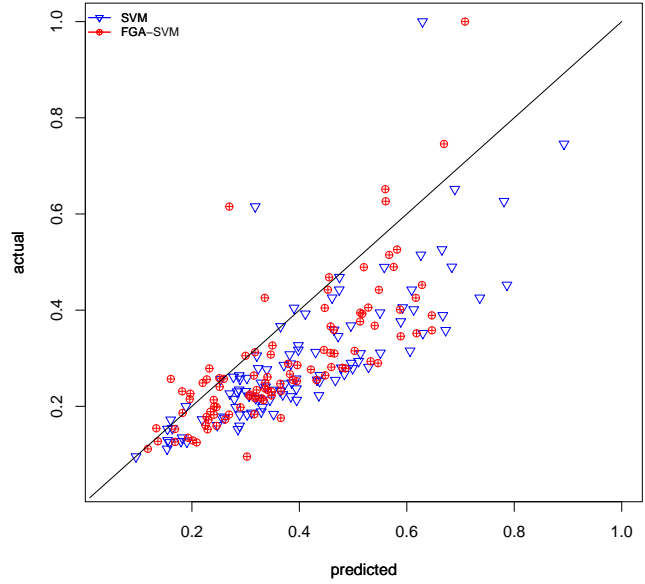}
\caption{Comparison showing predicted vs actual values on the test set for SVM and FGA-SVM.\label{svmfigs}}
\end{figure*}

\section{Stability analysis using incremental learning}
{\bf Theorem 2.}
The FGA has stability given by
\begin{align}
\beta_{\rm FGA} = \beta_{\rm SVM} + \sum_{v \in V} \beta_v \prod_{i_v=v}^{v_L} w_{i_v}
\end{align}
where $\beta_{\rm SVM}$ is the stability of a simple SVM, $\beta_v$ is the stability of an individual node $v$, and $w_{i_v}$ are the weights along paths from nodes $v \in \mathcal{V}$  to the root $v_L$ of the FGA tree.

{\bf Proof.} We can derive a relationship between the FGA and SVM stabilities as follows. Consider four cases
a) $SVM$ - no points removed
b) $SVM_{\backslash 1}$ - one point removed
c) $FGA$ - no points removed
d) $FGA_{\backslash1}$ - one point removed

The FGA starts off initialised to the simple SVM:
\begin{align}
FGA&=SVM\\
FGA_{\backslash 1}&=SVM_{\backslash 1}
\end{align}

Next, train the first node in the FGA (call this node $N_1$)
\begin{align}
FGA_{N1} &= SVM + N_1\\
FGA_{N_{1_\backslash 1}} &= SVM_{\backslash 1 } + N_{1\backslash 1}
\end{align}

The stability of the updated FGA is then
\begin{align}
\beta_{FGA_{N_1}} &= e-e' \\
&= FGA_{N_1} - FGA_{N_{1\backslash1}}\\
&= SVM - SVM_{\backslash 1} + N_1 - N_{1\backslash1}\\
&= \beta_{\rm SVM} + \beta_{N_1}.\prod_{i=1}^L w_i
\end{align}

Similarly for all other node retrainings, so
\begin{align}
\beta_{\rm FGA} = \beta_{\rm SVM} + \sum_{v \in \mathcal{V}} \beta_v \prod_{i_v=1}^L w_{i_v}
\end{align}

If we train only the first layer we are just adding beta terms for each node. We know that roughly,
\begin{align}
\beta_{\rm SVM} = \sum_{v_1 \in \mathcal{V}_1}  \beta_{v_1}
\end{align}
summed over first layer terms.

Thus typically the first layer variation will effectively give an FGA with double the SVM's beta. This result agrees well with observations, with FGA $\beta$ values between two and five times that of the simple SVM, with the ratio increasing with the number of layers.
$\Box$

\subsection{Stability results}
\begin{figure}[ht]
\includegraphics[width=7cm]{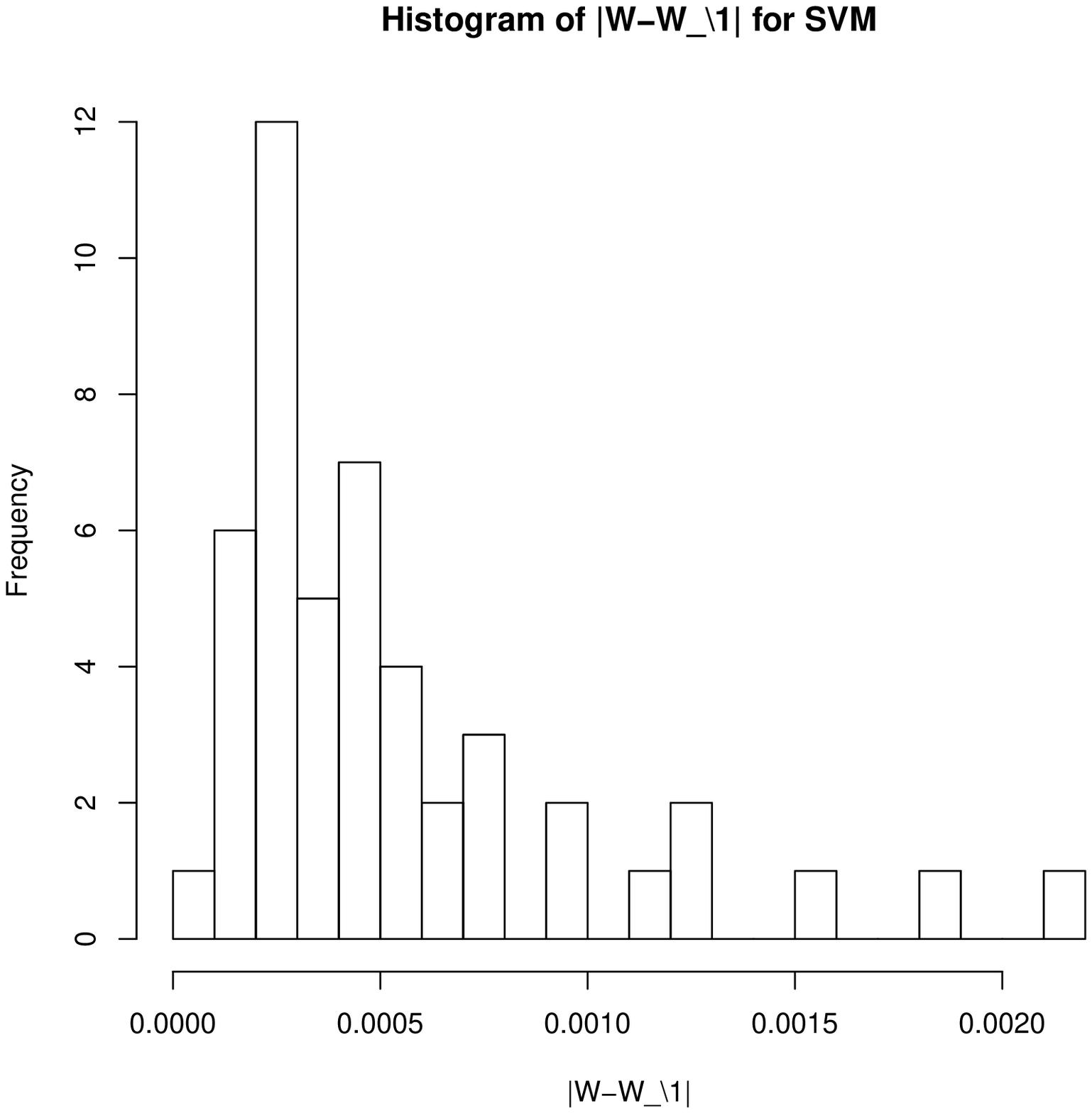}
\includegraphics[width=7cm]{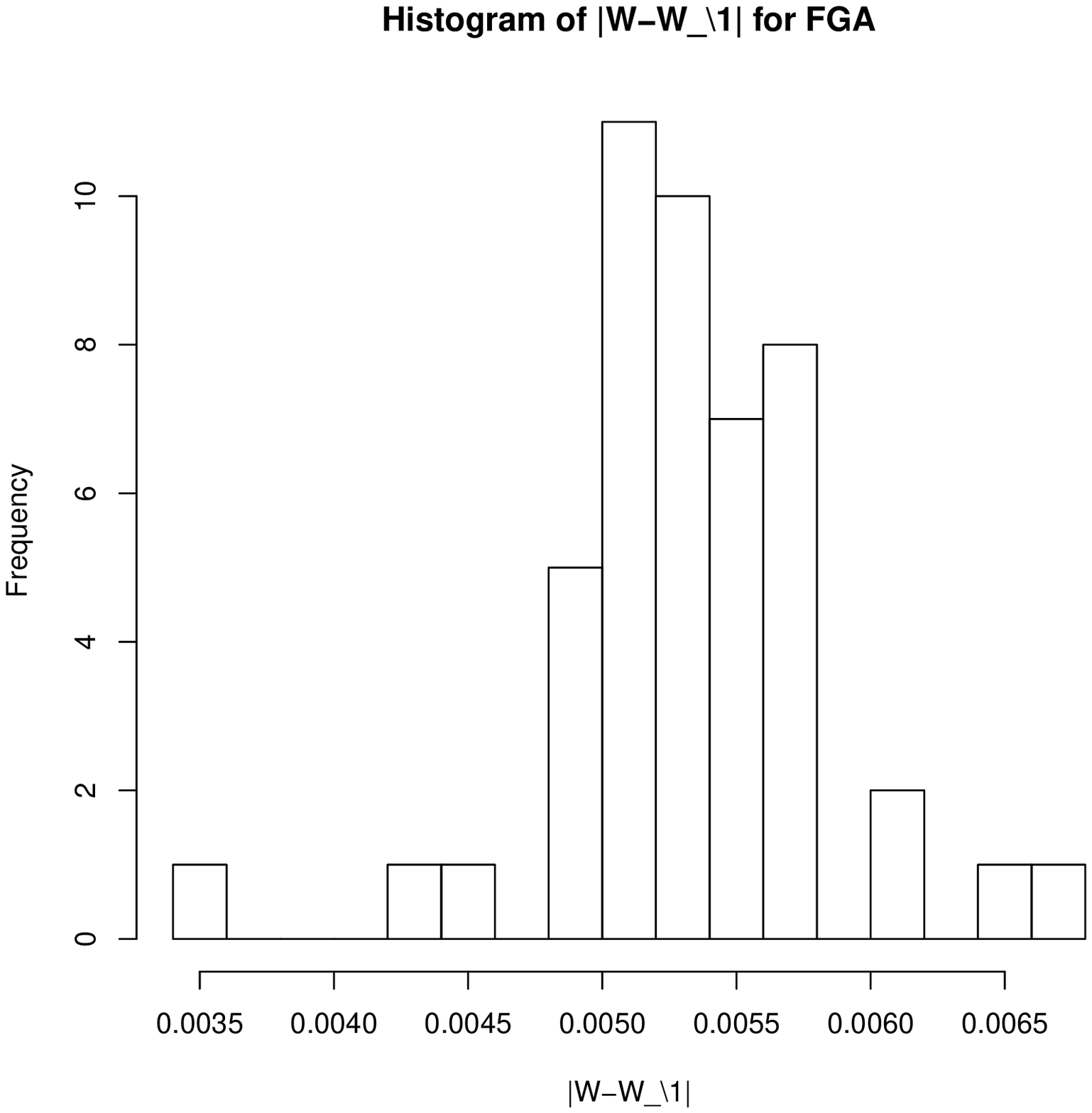}
\caption{Stability of weights \label{stabilityfigs} for the SVM and FGA. }
\end{figure}

\begin{figure}[ht]
\includegraphics[width=7cm]{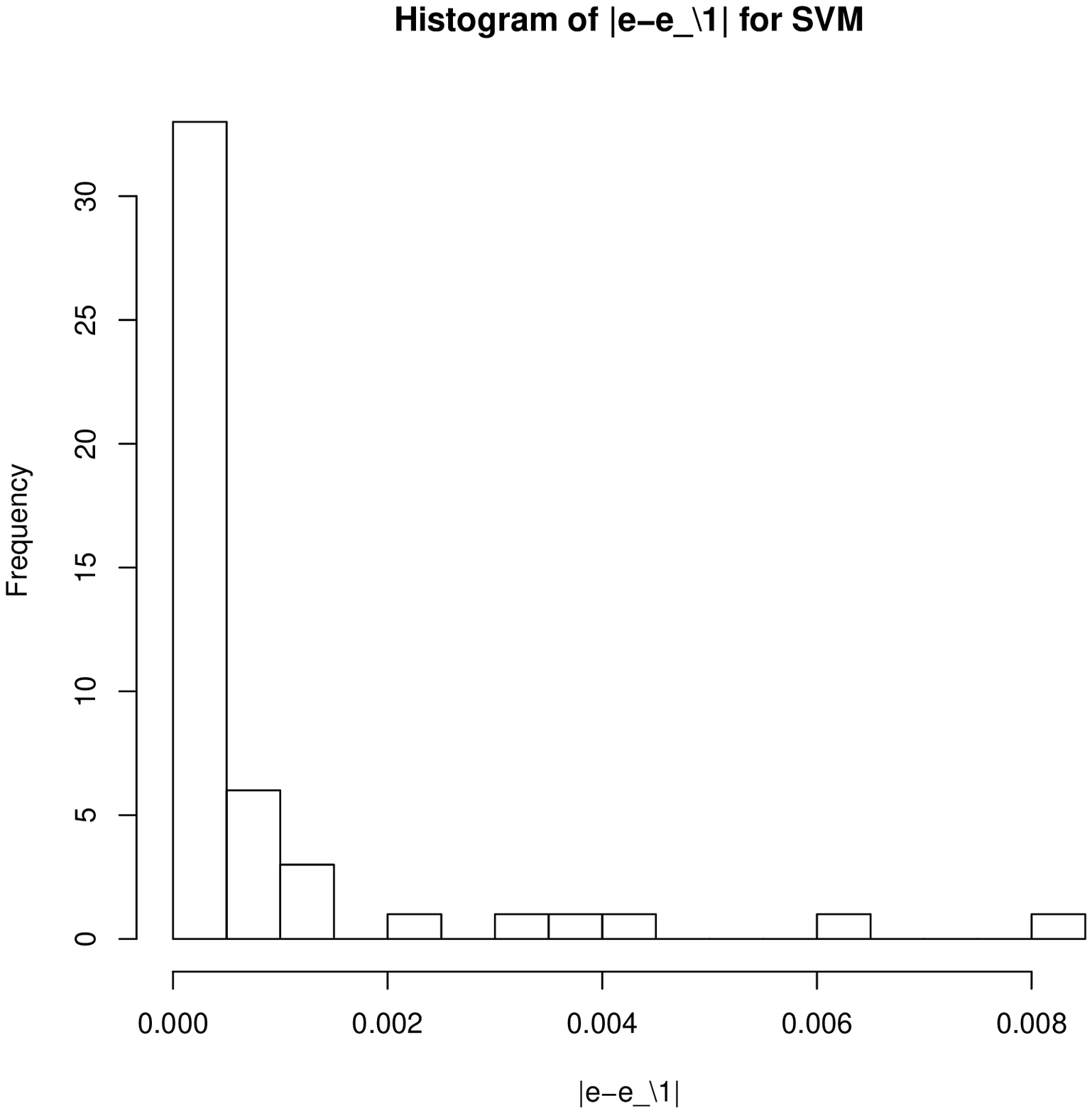}
\includegraphics[width=7cm]{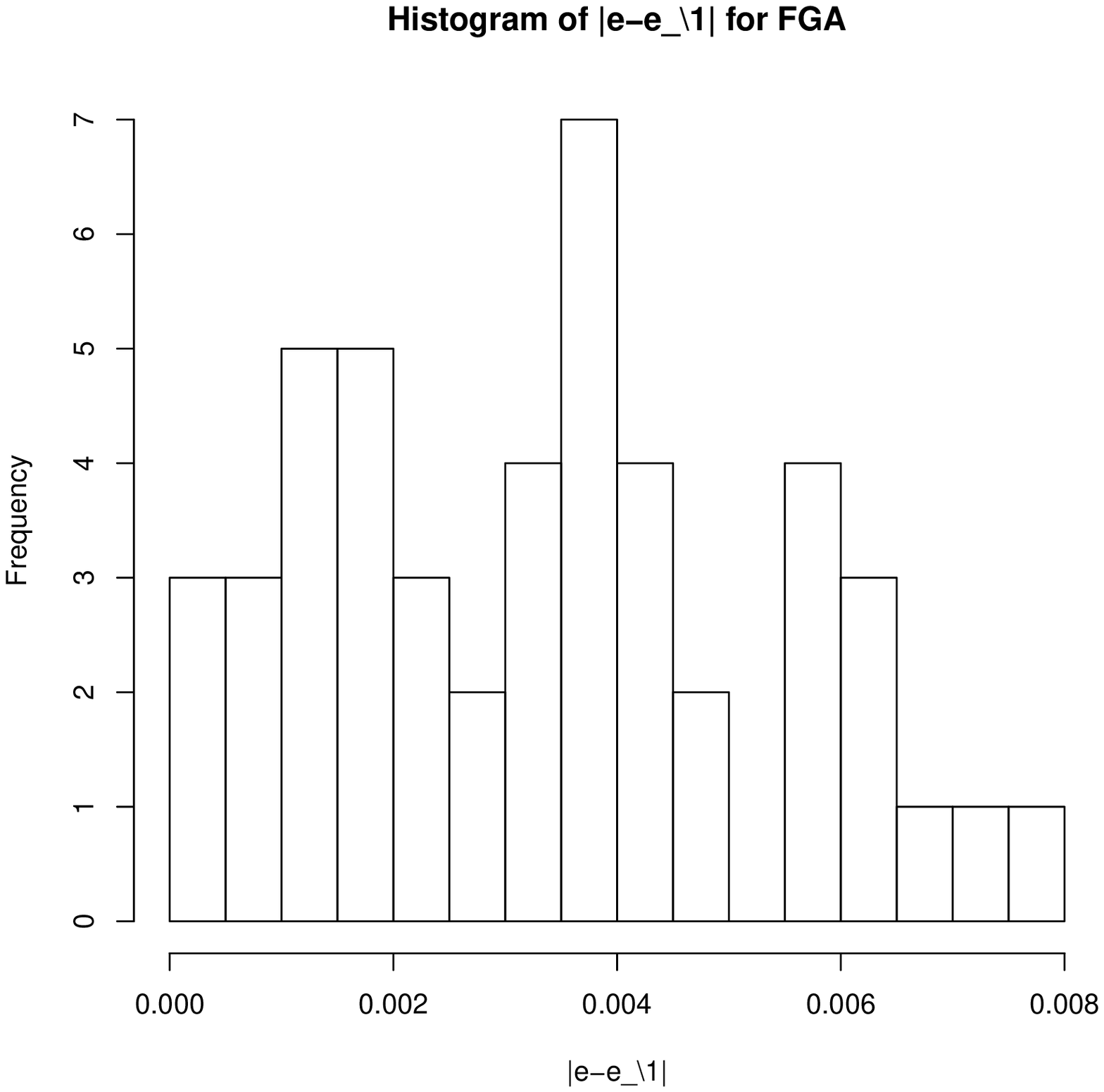}
\caption{Stability of test errors \label{stabilityerrorfigs} for the SVM and FGA. }
\end{figure}

Figure \ref{stabilityfigs} shows the changes in model weights as a result of removing one training point in a set of 50. We see the weights change by less for the SVM, but the FGA error is still relatively unchanged. This is a small price to pay for increased accuracy, shown in figure \ref{svmfigs}. In this example $w_{l=2}=1, m=5, L=2$ giving a beta ratio of $2$, which is close to the observed ratio in figure \ref{stabilityerrorfigs}.

Figure \ref{stabilityerrorfigs} shows the error distribution as a result of removing one training point in a set of 50. It is interesting to note that the range of the error differences is very similar, although the distribution is different.

\subsection{Stability as a function of the number of layers}
\begin{table}[!ht]
\centering
\begin{tabular}{ccc}
\hline\noalign{\smallskip}
L & m & $||e - e_{\backslash 1}||$ for 64 features  \\
\hline
\noalign{\smallskip}\hline\noalign{\smallskip}
6 & 2 & 0.02948 \\
4 & 4 & 0.02000 \\
3 & 8 & 0.01714\\
2 & 16 & 0.01441 \\
2 & 32 & 0.01740\\
1 & simple SVM & 0.009765\\
\hline
\end{tabular}
\caption{Stability of the FGA as a function of the number of layers for 64 features.\label{stability64}}
\end{table}

Tables \ref{stability64} and \ref{stability256} show the stability as a function of the number of layers and the number of features per node for 64 and 256 features respectively. The e1071 package \cite{e1071} was used for the SVM implementation. We see as the number of nodes and layers increases the stability of the FGA approaches roughly three times that of the SVM, in agreement with equation \ref{stabilityeqn}.

\begin{table}[!ht]
\centering
\begin{tabular}{ccc}
\hline\noalign{\smallskip}
L & m & $||e - e_{\backslash 1}||$ for 256 features  \\
\hline
\noalign{\smallskip}\hline\noalign{\smallskip}
8 & 2 & 0.009216 \\
4 & 4 &  0.009644 \\
3 & 8 & 0.008501\\
2 & 16 & 0.003544 \\
2 & 32 & 0.007390\\
1 & simple SVM & 0.001873\\
\hline
\end{tabular}
\caption{Stability of the FGA as a function of the number of layers for 256 features.\label{stability256}}
\end{table}

\section{Optimal Permutations}
{\bf Theorem 3.} Consider an FGA with four input features $1,2,3,4$. The second layer outputs are given by $X_{12},X_{34}$. A permutation of the input features gives second layer outputs $X_{13},X_{24}$. If the input vectors are equally weighted, in the sense that $K=\langle Y,X_{12}\rangle=\langle Y,X_{34}\rangle=\langle Y,X_{13}\rangle=\langle Y,X_{24}\rangle$
 and $\langle X_{12},X_{34}\rangle\leq \langle X_{13},X_{24}\rangle$
 then the empirical errors for the two permutations satisfy
\begin{equation*}
R_{12,34} < R_{13,24}
\end{equation*}

{\bf Proof.}
Here we are only considering the case where all vectors are normalised and have the same correlation with the target $Y$. Our assumption is that $\langle X_{12},X_{34}\rangle \geq \langle Y_{13},Y_{24}\rangle$.
All weights are uniform so the error is
\begin{align}
\langle Y - X_{12}& - X_{34},Y - X_{12} - X_{34}\rangle \\
=&\; \langle Y,Y\rangle +\langle X_{12},X_{12}\rangle +\langle X_{34},X_{34}\rangle \\
&- 2\langle Y,X_{12}\rangle- 2\langle Y,X_{34}\rangle+2\langle X_{12},X_{34}\rangle\\
\langle Y - X_{13}& - X_{24},Y - X_{13} - X_{24}\rangle \\
=&\; \langle Y,Y\rangle +\langle Y_{13},X_{13}\rangle +\langle X_{24},X_{24}\rangle \\
&- 2\langle Y,X_{13}\rangle- 2\langle Y,X_{24}\rangle+2\langle X_{13},X_{24}\rangle
\end{align}
thus
\begin{align}
R_{12,34}&= 1 +1 +1 - 4K +2\langle X_{12},X_{34}\rangle \\
&\leq R_{13,24}= 1 +1 +1 - 4K +2\langle X_{13},X_{24}\rangle
\end{align}
showing that minimising later-layer correlations leads to lower prediction errors for this simplified example.
$\Box$

As mentioned in the main article, the situation can become more complex. The table \ref{permutationttable} shows a search over multiple permutations, recording the best error found so far in the first column. The second column shows the number of statistically correlated adjacent variables in the first layer blocks of the FGA. The third column shows the sum of the p-values of these statistically correlated pairs. The fourth column is the average correlation of such pairs. We see there is no particular trend in columns three and four, but column two shows a clear trend: a better permutation is one with fewer correlations in the first layer. The reader may ask, isn't this in conflict with theorem 3? Theorem 3 assumes strong correlations in the higher order layers. The difference is that for this example, most of the second and higher order layers are not trained, so they are effectively bypassed. Only a few nodes in the whole FGA are trained, so it is a mostly shallow network. Thus it is preferable for those nodes to have fewer correlations in the early layers, and place the correlations in the few nodes that are retrained, which gives the FGA algorithm the best chance to model the difficult subspaces well.

\begin{table}[h]
\centering
\begin{tabular}{ccccc}
\hline
Permutation Improvement & Best Err & Sig p-value count  & Sum & Average  \\
\hline
1 &903.6  &      22 &0.0404 &0.00183\\
2 &815.6  &      19 &0.0362 &0.00191\\
3 &815.6  &      17 &0.0166 &0.00098\\
4 &769.1  &      15 &0.0151 &0.00101\\
5 &769.0  &      14 &0.0151 &0.00108\\
6 &769.0  &      14 &0.0151 &0.00108\\
7 &769.0  &      12 &0.0043 &0.00036\\
8 &567.1  &      11 &0.0333 &0.00303\\
\hline
\end{tabular}
   \caption{Successive permutation improvements. The p-values are summed over adjacent input features in the first layer of the FGA. as this FGA was quite flat, with only a few retrainings in the second and higher layers, it is better to have less correlation in the early layers. \label{permutationttable}}
\end{table}

\end{document}